\documentclass[letterpaper, 10 pt]{IEEEtran}
\IEEEoverridecommandlockouts
\usepackage{filecontents}
\usepackage[space,sort]{cite}
\usepackage{amsmath}
\usepackage{amsfonts}
\usepackage{amsthm}
\usepackage{dsfont}
\usepackage{amssymb}
\usepackage[ruled,vlined]{algorithm2e}
\usepackage{mathtools}
\usepackage{tabularx}
\usepackage{graphicx}
\usepackage{subfigure}
\usepackage{enumerate}
\usepackage{float}
\usepackage{url}
\usepackage{verbatim}
\usepackage{booktabs} 
\usepackage{multirow}
\usepackage[linkcolor=black,citecolor=black,urlcolor=black,colorlinks=true]{hyperref}
\usepackage[dvipsnames]{xcolor}
\usepackage[most]{tcolorbox}
\usepackage[inline]{enumitem}
\usepackage[normalem]{ulem}
\usepackage{bm}
\graphicspath{{figures/}}

\DeclareMathOperator*{\argmin}{arg\,min}

\usepackage{algpseudocode}

\newtheorem{problem}{Problem}
 \newcommand{\add}[1]{{{{\color{black}#1}}}}
\newcommand{\delete}[1]{{\bgroup\markoverwith{\textcolor{red}{\rule[0.5ex]{2pt}{0.4pt}}}\ULon{#1}}}

\newcommand{\tabincell}[2]{\begin{tabular}{@{}#1@{}}#2\end{tabular}}

\begin{document}
\title{Deep Learning for Optimization of Trajectories for Quadrotors}
	\author{Yuwei Wu, Xiatao Sun, Igor Spasojevic, Vijay Kumar
	\thanks{The authors are with the GRASP Laboratory, University of Pennsylvania, Philadelphia, PA, 19104 USA {\tt\small\{yuweiwu, sxt, igorspas, kumar\}@seas.upenn.edu}.}}

\maketitle
\thispagestyle{empty}
\pagestyle{empty}
\begin{abstract}

This paper presents a novel learning-based trajectory planning framework for quadrotors that combines model-based optimization techniques with deep learning. Specifically, we formulate the trajectory optimization problem as a quadratic programming (QP) problem with dynamic and collision-free constraints using piecewise trajectory segments through safe flight corridors \cite{7839930}. 
We train neural networks to directly learn the time allocation for each segment to generate optimal smooth and fast trajectories.  
Furthermore, the constrained optimization problem is applied as a separate implicit layer for \add{backpropagation} in the network, for which the differential loss function can be obtained.
We introduce an additional penalty function to penalize time allocations which result in solutions that violate the constraints to accelerate the training process and increase the success rate of the original optimization problem.
To this end, we enable a flexible number of sequences of piece-wise trajectories by adding an extra end-of-sentence token during training. We illustrate the performance of the proposed method via extensive simulation and experimentation and show that it works in real time in diverse, cluttered environments.

\end{abstract}

\section{Introduction}
	
The application of micro aerial vehicles (MAVs) has significantly expanded recently in agriculture, industrial inspection, and customer services \cite{10160295}. 
As one crucial module of autonomy, the research of trajectory generation has made significant progress in realistic feasibility and computation efficiency. Traditional optimization-based trajectory generation with simplified dynamics offers computational advantages for real-time deployments \cite{5980409} but suffers from parameter tunings and infeasible formulations when applied to real-world data. Moreover, existing learning-based methods \cite{9794627, 6630809, doi:10.1126/scirobotics.abg5810} are mainly trained to generate commands or waypoints, limited by dynamic interpretability and fail to provide guarantees for real-time safe flight. This paper addresses online trajectory generation, focusing on achieving efficiency, reliability, and optimality.

\begin{figure}[t]
      \centering
      \includegraphics[width=1\columnwidth]{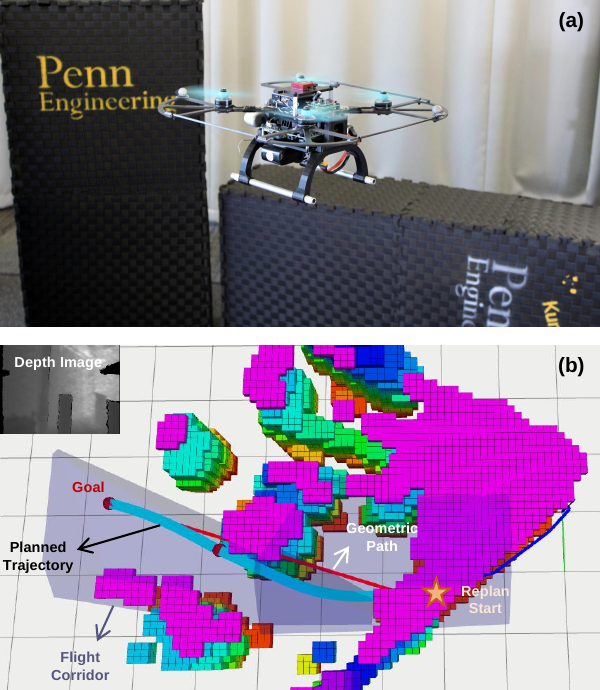}
      \caption{(a) Real-time deployment on Falcon 250 v2 autonomous stack. (b) The planned trajectory (in light blue) demonstration with time allocation. We use a safe flight corridor (in purple) and a geometrical searched path (in red) and allocate time by a neural network. Then the trajectory generation is solved by fixed-time \add{quadratic} programming. The video is available at: \url{https://youtu.be/tA02dJz9ux8}}
      \label{fig: fig1}
      \vspace{-0.2cm}
\end{figure}
There is extensive work on optimization-based approaches that rely on approximating the free space by safe-flight corridors which are defined by a set of overlapping convex subsets. 
These approaches find the optimal trajectory by solving a quadratic program that assumes a specified {\it time allocation}, an assignment of time intervals to each convex subset \cite{5980409}. However, there are no principled approaches to solving the time allocation problem. Most previous work decouples the time allocation problem from the optimization problem and \add{focuses} on refining and scaling the time with heuristics to arrive at a feasible time allocation\add{\cite{7487283, richter2016polynomial, tordesillas2021faster, 9197157, 9765821, doi:10.1177/0278364914558129, 8593579, 9147300, doi:10.1126/scirobotics.abh1221}.} Parameterizing the optimal time is computationally expensive, and even so, does not provide optimality guarantees. Including time parametrization results in a joint optimization of piecewise trajectories and time intervals, leading to suboptimality. Most state-of-the-art solvers cannot find feasible solutions with a reasonable computational budget.
These works inspired us to explore the nature of trajectory generation as an optimization problem and improve the efficiency and optimality of the existing methods. 

Leveraging the advantages of both model-based and learning-based methods, we propose an efficient learning-based planning framework that utilizes neural networks to solve the optimal time allocation problem for each minimum control trajectory. 
The trajectory generation problem is formulated by finding the optimal trajectory coefficients with discrete dynamic constraints and loose constraints for intermediate waypoints in overlapping regions of flight corridors. 
We present a neural network to output time allocation, with an extra differential implicit layer \cite{10.5555/3305381.3305396} solving trajectory optimization with hard constraints parameterized by time allocations. 
In addition, we design loss functions related to the original nonlinear programming while considering different solving results and adding infeasible penalty terms to train the network efficiently.
We summarize our contributions as follows:

\begin{itemize}
    \item A neural network architecture incorporates trajectory optimization problems as implicit layers with a flexible sequence of trajectory segments.
    \item An interpretable and lightweight learning-based planning framework that directly learns optimal time allocation from the corridor and state constraints for minimum control trajectory generations with quadratic programming.
    \item Real-world deployment of our proposed method on hardware platform integrating on-board autonomy and high-performance demonstration in both indoor and outdoor environments, shown in Fig. \ref{fig: fig1}. The open source code is available at: \url{https://github.com/KumarRobotics/AllocNet}.
\end{itemize}

\section{Related Works}
\label{sec:related_works}

\subsection{Optimal Time Allocation for Trajectory Planning} 

Decoupled trajectory parameterization of flat outputs with polynomials has been well-explored as an efficient way to optimize the trajectory with quadratic programming (QP) while encoding dynamic and obstacle-free constraints with differential flat outputs  ~\cite{5980409, doi:10.1177/0278364914558129}. 
\add{A closed-form solution for this QP formulation, incorporating equality constraints, was proposed in \cite{doi:10.1177/0278364914558129}. Its numerical stability issues during matrix inverse were addressed in \cite{7963135}, and better-conditioned matrices are formulated in \cite{9341794} with linear computational complexity to solve QP.}
However, introducing time allocation optimization for piece-wise trajectory makes the problem more challenging, resulting in a joint spatial-temporal ill-conditioned nonlinear programming (NLP). 
Previous research has primarily focused on iterative time refinement~\cite{7353622}, \add{time-optimal primitives \cite{8206119, penicka2022minimum}, path parameterization~\cite{doi:10.1177/027836498500400301, 8593579}}, or bi-level optimization~\cite{9147300} to address the challenge by solving the problem in two phases or iteratively searching for optimal time allocation. Moreover, different analytic gradients towards time and waypoints have been studied~\cite{9765821} to improve the efficiency of fast trajectory generation. However, the lack of feasibility guarantees and the high sensitivity of parameters limit the performance of these approaches in real-world scenarios.

\subsection{Learning for Autonomous \add{Navigation}} 

Learning-based methods have emerged as powerful approaches for autonomous \add{navigation} applications. 
End-to-end policy learning~\cite{9794627} directly trained the control commands using deep reinforcement learning to generate trajectory with waypoints. 
Many existing works assume a collision-free reference line and perform policy learning based on state observations along this predefined path or in an obstacle-free space. 
These methods involve imitating human drivers or expert planners \cite{6630809, doi:10.1126/scirobotics.abg5810} or utilizing learning-based perception to generate collision-free waypoints \cite{pmlr-v100-bansal20a}. 
However, they often generate \add{over-simplified} end-to-end commands or waypoints that require additional components such as Model Predictive Control (MPC) or a dynamic trajectory planner to obtain a smoother trajectory \cite{46570, 8593536, yang2023iplanner}. \add{Differentiable trajectory optimization mapping from waypoints can also be included during the learning process \cite{yang2023iplanner}}.
\add{However}, the computational complexity associated with the learning processes and generalization limits the deployment to more unknown scenarios.
Addressing time allocation in trajectory optimization, Ryou et al. \cite{pmlr-v205-ryou23a} employed a deep neural network with sequence-to-sequence learning to determine optimal time allocation given waypoints. Nevertheless, the complex process of multi-phase learning with brute-search time allocations remains a challenge that must be addressed, encouraging us to pursue a lightweight and efficient structure.

\subsection{Neural Optimizations with Hard Constraints} 

Recent works using neural networks (NNs) to learn optimizers with hard constraints have gained significant attention due to \add{their} extensive applications and promising results. Amos et al. \cite{10.5555/3305381.3305396, agrawal2019differentiable} proposed OptNet to demonstrate the effective integration of quadratic programming (QP) with hard constraints within the NNs framework, which illustrate the potential of leveraging implicit layers to tackle complex optimization problems. Jaquier et al. \cite{jaquier_learning_2022} extended the application of OptNet to multitask control problems by formulating these tasks as QPs and employing differentiable layers to enable the learning of specific manipulator skills. However, these approaches primarily focus on feasible problem formulations, leaving a gap when dealing with the inherent \add{infeasibility} of specific optimization problems. To address this limitation, DC3~\cite{donti_dc3_2021} fixed the violation and \add{obtained} an approximated solution by adding additional differentiable completion and correction procedures, which can be extended to nonlinear optimization. Other works using meta-learning \cite{negiar_learning_2023}, double descriptions \cite{9150971} can also be applied to learn optimizers. Integrating differential optimization in NNs provides diverse tools to address optimization problems. 

\section{Deep Time-optimal Trajectory Learning}
\label{sec:deep_time_allocation_and_trajectory_learning}

\subsection{Spatial-temporal Trajectory Optimization}

Leveraging the differential flatness of quadrotor dynamics, we represent the trajectory of the quadrotor by the trajectory of its flat outputs: $\bm{\sigma} = [x, y, z, \psi]^T \in \mathbb{R}^3 \times S^1$. 
The vector $[x, y, z]^T \in \mathbb{R}^3$ represents the position of the quadrotor, whereas $\psi \in S^1$ represents the ``yaw'' angle that the projection of its body $x$ axis forms with the $x$ axis of the world frame. 
The remarkable fact is that a generic trajectory of the quadrotor corresponds to a unique function $\bm{\sigma}$, and in turn, every sufficiently smooth $\bm{\sigma}$ corresponds to a dynamically feasible trajectory of the vehicle, \textit{provided its motors can exert sufficient thrust.} 
The latter\add{, actuation constraints} render numerous classes of trajectory optimization problems for quadrotors computationally challenging. 
In this work, we focus on the class of minimum-time problems, enforcing the constraints in an approximate discrete approach on the derivatives of $\bm{\sigma}$, in the hope of developing an algorithm that can output near-optimal and \textit{safe} (collision-free, dynamically feasible) trajectories in real-time. 
\begin{figure}[!ht]
      \centering
      \includegraphics[width=1\columnwidth]{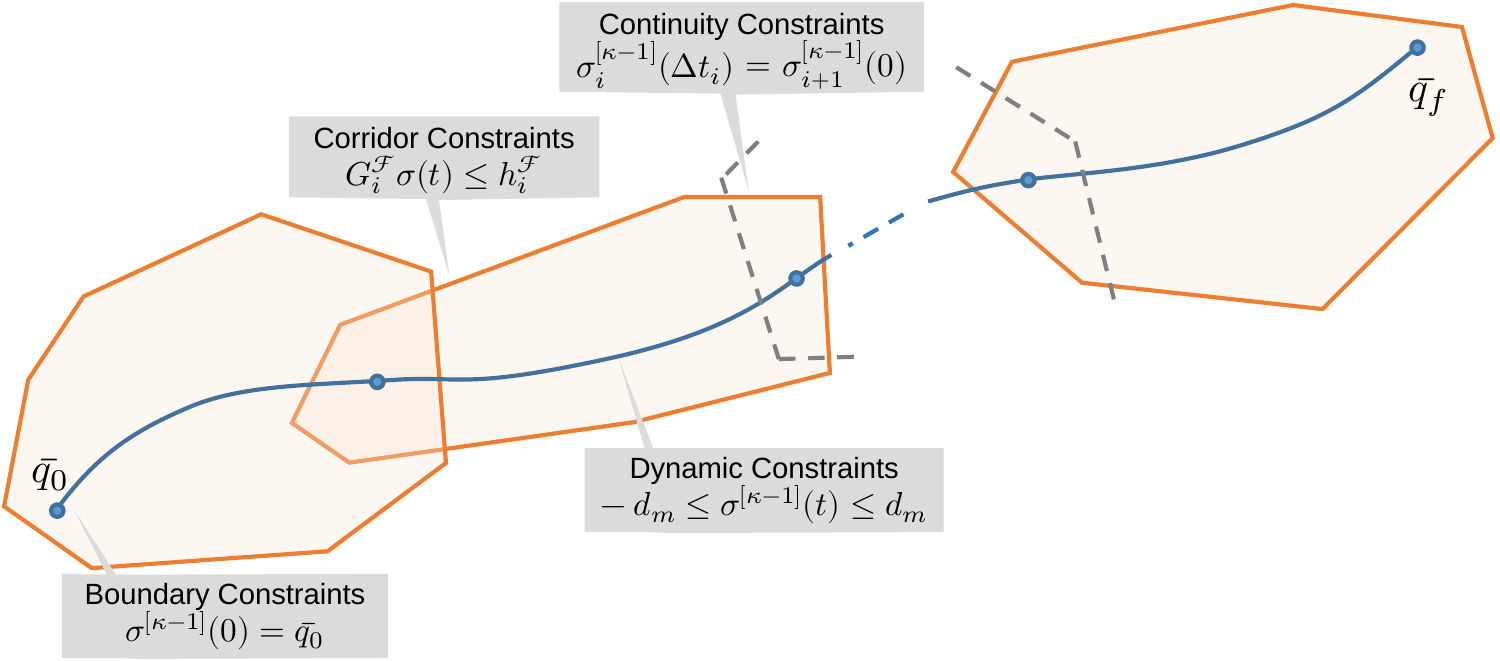}
      \caption{Trajectory Optimization formulation with corridors and dynamic constraints. The equality boundary and continuity constraints are also demonstrated.}
      \label{fig: traj_opt}
\end{figure}
We parameterize the trajectory of flat outputs $\bm{\sigma}(t)$ as a suitably smooth $N = 2\kappa - 1 $ degree piece-wise polynomial function.
In particular, $\bm{\sigma}(t)$ is a concatenation of a sequence of $M \in \mathbb{N} $ segments, $\bm{\sigma}_{1}(\cdot), \bm{\sigma}_{2}(\cdot), ..., \bm{\sigma}_{M}(\cdot)$. 
Segment $i$, $\bm{\sigma}_{i}$, is characterized by its coefficients $\mathbf{c}_{i} \in \mathbb{R}^{(N+1) \times 3}$ and duration $\Delta t_i > 0$.
The complete trajectory $\bm{\sigma}(t)$ is therefore encoded through its coefficient matrix $\bm{c} = [ \bm{c}_1^T, ..., \bm{c}_M^T   ]^T$ and time allocation intervals $\bm{t} = [ \Delta t_1, ...,  \Delta t_M]^T$, with:
\begin{equation}
\begin{aligned}
\bm{\sigma}(t) =   \bm{\sigma}_i(t - & \sum_{k = 1}^{i-1} \Delta t_{k} ) ,  \ \forall  t \in [\sum_{k = 1}^{i-1} \Delta t_{k} , \sum_{k = 1}^{i} \Delta t_{k}],  \\
\mathrm{where} \quad  \bm{\sigma}_i(t) &= \bm{{\rm c}}_i^T \bm{\beta}(t),\\
     \bm{\beta}(t) &= [1, t, t^2, ..., t^N]^T, \  \forall  t \in [0, \Delta t_i].
\end{aligned}
\end{equation}
Given a sequence of safe flight corridors extracted from free space, together with dynamic feasibility in the form of collision-free constraints on the derivatives of the flat outputs, generate the time-optimal vector of duration intervals and corresponding coefficients. 
For example, ``collision-free'' constraints on the second derivatives encode actuation bounds on the acceleration of the vehicle, constraints on the first derivatives encode bounds on its velocity, and constraints on the zeroth derivatives encode staying in the obstacle-free region of space, shown in Fig. \ref{fig: traj_opt}. 

The objective function involves minimizing the traversal time as well as the control effort (snap or jerk) required to execute the planned $n$-dimension piece-wise polynomial trajectory. Hence, the problem is formulated by
\begin{subequations}
        \begin{align}
	\label{eq:optimization}
	\min_{\bm{c, t}} & \int_{0}^{\sum_{k = 1}^{M} \Delta t_{k}} \| \bm{\sigma}^{(\kappa)}(t) \|^2 dt + w_t  f(\bm{t}) , \\
	\ {\rm s.t.} \ \ 		
	&\ \  \bm{\sigma}^{[\kappa-1]}(0) = \bar{\bm{q}}_0, \ \bm{\sigma}^{[\kappa-1]}(\sum_{k = 1}^{M} \Delta t_{k}) = \bar{\bm{q}}_f,  \\
 	&\ \ \bm{\sigma}_i^{[\kappa-1]}(\Delta t_i) = \bm{\sigma}_i^{[\kappa-1]}(0), \\
	& \  - \bm{d}_{m} \leq  \bm{\sigma}_i(t)^{[\kappa-1]} \leq  \bm{d}_{m} , \label{ineq: dynamic} \\
	&\  \ \bm{G}_{i}^{\mathcal{H}}\bm{\sigma}_i(t) \leq \bm{h}_{i}^{\mathcal{H}} ,\quad \label{ineq: corridor} \\
     &\ \ \ \forall i \in {1, ... , M}, \ \ \forall  t \in [0, \Delta t_i] .
        \end{align}
\end{subequations}
The $\bar{\bm{q}}_0,  \bar{\bm{q}}_f \in \mathbb{R}^ {3 \times \kappa} $ are initial and end states up to ($\kappa-1$)-order derivative and $\bm{\sigma}^{[\kappa-1]}(t) = [  \bm{\sigma}(t)^T, ... ,   \bm{\sigma}^{(\kappa-1)}(t)^T]^T $. 
The time objective $f(\bm{t})$ could be the sum of total traversal time or its absolute deviation from an expected finish time, and $w_t$ is the associated weight.
We use Eq. (\ref{ineq: dynamic}) to include all state and input inequality constraints except the constraints on position, which are captured in Eq. (\ref{ineq: corridor}).
The vector $\bm{d}_{m} \in  \mathbb{R}^{\kappa-1}$ includes the magnitude of maximum velocity $v_{max} $ and acceleration $a_{max}$ for $k=3$, adding maximum jerk $j_{max} $ for $k=4$. 
The corridor constraints are formulated by half-space representation of polytopes with $\bm{G}_{i}^{\mathcal{H}} \in \mathbb{R}^{F_i \times 3}$, $\bm{h}_{i}^{\mathcal{H}} \in \mathbb{R}^{F_i} $, where $F_i$ is the number of hyperplanes for $i$-th polytope. 
The continuous inequality constraints are enforced by discretizing the trajectory by the sampling number $N_{res}$ and querying sample points to satisfy positions within corridors and higher derivatives within maximum limits. 
More generally, we can formulate the problem into nonlinear programming as follows.
\begin{problem}
\label{prob: nonlinear}
The Time-optimal minimum control trajectory optimization with corridor and dynamics constraints is a parametric nonlinear programming (NLP) problem.
\begin{subequations}
        \begin{align}
        \min_{ \bm{c, t}} \ \bm{c}^T \bm{Q}(\bm{t}) \bm{c} &  + w_t  f(\bm{t}) , \\
        {\rm s.t.}   \quad  \bm{A}(\bm{t}) \bm{c} & = \bm{b}, \\
         \bm{G}(\bm{t}) \bm{c} & \leq \bm{h}, \\
          \qquad \quad  \ \ \bm{H t} & = \bm{m}, \\
        \quad  \bm{C t} & \leq \bm{d}, 
        \end{align}
\end{subequations}
where $\bm{Q} \succeq  \bm{0} $ is a positive semidefinite matrix. 
\end{problem}

The former problem presents a non-convex optimization that nevertheless possesses a certain amount of structure. 
In particular, given a feasible time allocation, we can directly compute the time-dependent matrices $\bm{Q_t}, \bm{A_t}, \bm{G_t}$ and solve for the optimal vector of coefficients as a convex \textit{quadratic} programming.
Importantly, all aforementioned constraints are \textit{linear} constraints on $\mathbf{c}$ for a fixed $\mathbf{t}$.


\begin{problem}
\label{prob: traj_opt}
The minimum control trajectory optimization for a given time allocation can be formulated as a quadratic programming (QP) problem as follows:
\begin{subequations}
        \begin{align}
     \min_{\bm{c}} \  \bm{c}^T  & \bm{Q_t} \bm{c} , \\
     \quad {\rm s.t.} \    \ \bm{A_t} \bm{c}  &= \bm{b},   \\ \bm{G_t} \bm{c}   &\leq \bm{h}.
        \end{align}
\end{subequations}
\end{problem}
Unfortunately, the \textit{quadratic} problem is not always feasible with corridors and dynamics constraints for a specific time allocation, which introduces extra refinements and iterations for better time intervals.

\subsection{Learning Time Allocation with NNs}
\begin{figure*}[!ht]
      \centering
\includegraphics[width=2.02\columnwidth]{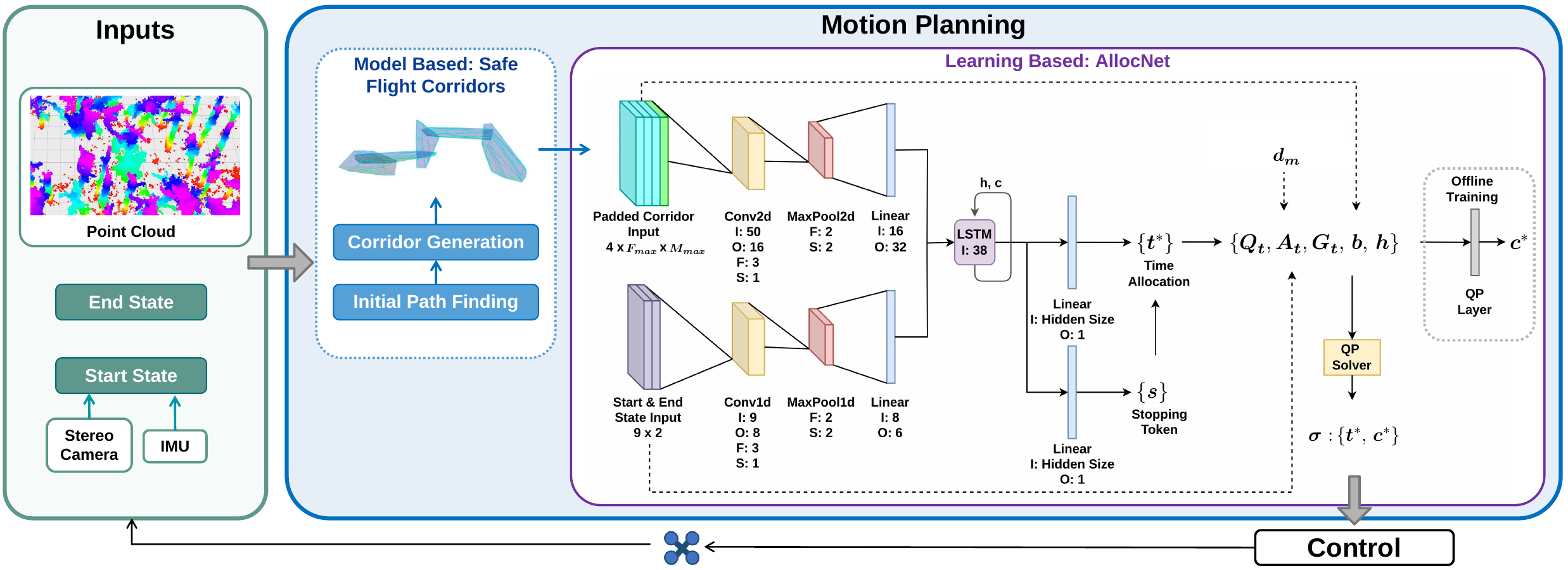}
      \caption{The overall planning framework. The network architecture is included in the motion planning part. With some initial path generation and corridor generation, we can use two terms of network input: padded corridor and start and end state input. ``I", ``O" represent the input and output size, ``F" represents kernel size, and ``S" is stride. The output layer for time allocation is connected with a Softplus function to ensure $\bm{t}^* \succ \bm{0}$. During the offline training, we utilize a QP implicit layer to propagate gradients toward time allocation, and for online planning, we directly use a QP solver to solve the minimum control trajectory optimization.}
      \label{fig: network}
\end{figure*}

\subsubsection{Network architecture}

We propose a Time Allocation Network (\textit{AllocNet}) to learn the time allocation for trajectory generation.
The full network architecture and system framework are shown in Fig. \ref{fig: network}.
We learn a function that maps the flattened initial and terminal states, denoted by $\bm{\mathrm{q}} = [ \bar{\bm{\mathrm{q}}}_0, \bar{\bm{\mathrm{q}}}_f]^T$, together with the sequence of feasible flight corridors, denoted by  $  \{ \mathcal{P}^{\mathcal{H}}_i  = \left [ \bm{G}_{i}^{\mathcal{H}}, \bm{h}_{i}^{\mathcal{H}} \right ]^T  \}_{i=1}^{M_{max}}  $, to the sequence of optimal times, $\bm{t}^{*} \in \mathbb{R}^M$.
Each flight corridor is represented as one matrix in $\mathbb{R}^{F_{max} \times 4}$. 
Here $F_{max}$ represents the maximum number of faces of the corridor, that is, the number of linear inequality constraints that define it, and $4$ corresponds to the encoding of each such inequality using the unit normal and the signed distance of the hyperplane that represents its zeroth level set from the origin. 
We note that the actual number of polytopes $M$ need not be fixed; it can take on any value at most $M_{max} $.
At the level of inputs to  \textit{AllocNet}, this is handled by zeroing out the data corresponding to the trailing $\mathcal{P}_i^{\mathcal{H}}$.
We further introduce another, auxiliary, output of the network - a ``stopping token'' \cite{10.5555/2969033.2969173}  that signals the maximum index of the $M$-dimensional output vector $\bm{t}$ that corresponds to the traversal time of a supplied corridor.


\subsubsection{Implicit differentiable layers}

With time allocation from the previous layer, we can explicitly formulate the trajectory optimization problem as a quadratic programming problem. The matrices formed by time allocation are defined as a set $\Phi(\bm{t}) = \{\bm{Q}(\bm{t}), \bm{A}(\bm{t}), \bm{G}(\bm{t}) \}$. Then we introduce a differentiable layer with its input and output as:

\begin{subequations}
        \begin{align}
        \bm{z}_{i+1} = \argmin_{ \bm{z}} \  \bm{z}^T  & \bm{Q}(\bm{z}_i)  \bm{z},  \\
        {\rm s.t.}   \quad  \bm{A}(\bm{z}_i) \bm{z} & = \bm{b}, \\
         \bm{G}(\bm{z}_i) \bm{z} & \leq \bm{h},
        \end{align}
\end{subequations}
where $\bm{z_i} = \bm{t}^*$ \add{refers to the optimal time allocation from the previous layer}, $\bm{z_{i+1}} = \bm{c}^*$ \add{is the optimal coefficient matrix}. After solving this problem, we can get the equalities part of the KKT condition \add{\cite{boyd2004convex}} with the primal-dual optimal solution,
\begin{equation}
\Gamma(\bm{c}^*, \nu^*, \lambda^*, \Phi) = \begin{bmatrix}
\ 2 \bm{Q} \bm{c}^* +  \bm{A}^T \nu^* +  \bm{G}^T \lambda^*  \\ 
\   \bm{A} \bm{c}^* -  \bm{b}  \\
 \mathrm{diag}  ( \lambda^*) ( \bm{G} \bm{c}^* -  \bm{h})
\end{bmatrix} = 0.
\end{equation}

To keep it short, we use $y^* = (\bm{c}^*, \nu^*, \lambda^*)$ in the rest of paper. By using implicit differentiation, we can get
\begin{equation}
\frac{ \partial  \Gamma(y^*, \Phi)}{\partial 
 \Phi } + \frac{ \partial  \Gamma(y^*, \Phi) }{\partial y^* }  \frac{ \partial y^*}{\partial 
 \Phi} =0.
\end{equation}
Therefore, we can expand the partial derivative of output towards input as time allocation with chain rules by:
\begin{equation}
\label{eq:jaco_time}
\frac{ \partial y^*}{\partial 
\bm{t}^*}  = \frac{ \partial y^*}{\partial 
 \Phi}  \frac{ \partial \Phi }{\partial 
 \bm{t}^*} 
 = - \left (\frac{ \partial  \Gamma(y^*, \Phi) }{\partial y^* } \right )^{-1} \frac{ \partial  \Gamma(y^*, \Phi)}{\partial 
 \Phi } \frac{ \partial \Phi }{\partial 
 \bm{t}^*}.
\end{equation}
With the above expression, we are able to \add{backpropagate} gradients of the primal solution $\bm{c}^*$ towards time $\bm{t}^*$.

\subsubsection{Loss functions}

 We consider two main loss functions, one loss term $\ell_F $ for min objective and a stopping token loss $\ell_T$ for a flexible number of segments. The loss function is evaluated by the objective of the original nonlinear programming
\begin{equation}
\ell_F(\bm{t}^*) =  (\bm{c}^*)^T  Q(\bm{t}^*) \bm{c}^* + w_t \bm{\mathrm{1}}^T \bm{t}^*,
\end{equation}
where $\bm{\mathrm{1}}$ is an \add{ vector of ones} to minimize the total time term. 
We need to propagate the gradient toward the previous layer. Based on Eq. (\ref{eq:jaco_time}), we can further utilize chain rules to get all gradients to time allocation, as:

\begin{equation}
\triangledown_{t} \ell_F(\bm{t}^*) = \left (\frac{\partial Q}{ \partial \bm{t}^* } \right )^T \triangledown_{Q} \ell_F + \left (\frac{\partial \bm{c}^*}{ \partial \bm{t}^* } \right )^T \triangledown_{\bm{c}^*} \ell_F + w_t \bm{\mathrm{1}}
\end{equation}

\add{The QP implicit layer assumes the feasibility of the problem and the satisfaction of the KKT condition. However, there are no guarantees to formulate problems that are always feasible primarily because of invalid time allocations. Additionally, in certain scenarios, the QP solver cannot find an optimal solution within the allowed time constraints.} 
In differentiable optimization layers with learning hard constraints, usually, the relaxed problems are studied to ensure feasibility, which will leave the gap to the original problem \add{\cite{10.5555/3305381.3305396}}. 
For the conditions of solver failures, we instead use a different loss term to penalize the deviation to the reference feasible time allocation,
\begin{equation}
\ell_{F}(\bm{t}^*) =  w_F \| \bar{\bm{t}}- \bm{t}^* \|^2 + w_t \bm{\mathrm{1}}^T \bm{t}^*,
\end{equation}
where $w_F$ is the weight for the MSE loss. 

When generating the overlapping flight corridors, the number of segments usually depends on the initial seed path for polytopes and goal position, which we expect to represent the problem efficiently. To maintain such an efficient representation, a flexible segment length rather than a redundant corridor sequence would be much better for formulating the optimization. 
Thus, we design the stopping token loss for \textit{AllocNet} that considers the timeliness of terminating the sequence. 

More specifically, the token loss function aims to minimize three terms: the prediction error, the cost associated with prematurely ending the sequence, and the cost associated with ending the sequence late. The token loss function $\ell_{T}$ is formulated as follows:
\begin{equation}
\ell_{T}(\bm{s}) = \ell_{BCE}(\bm{s}, \bar{\bm{s}}) + \lambda_{p} \cdot (\ell_{PEP} + \ell_{LEP}),
\end{equation}
where $\bm{s} = [s_1, ..., s_{M_{max}}]$ represents the predicted stop tokens, $ \bar{\bm{s}} = [\bar{\bm{s}}_1, ..., \bar{\bm{s}}_{M_{max}}]$ is the ground truth stop tokens, $\ell_{BCE}$ is the Binary Cross Entropy loss, $\ell_{PEP}$ is the premature end penalty, $\ell_{LEP}$ is the late end penalty, and $\lambda_{p}$ is the penalty weight for the premature and late end penalties. We can define the premature end penalty $\ell_{PEP}$ and late end penalty $\ell_{LEP}$ as:

\begin{equation}
\ell_{PEP} = \sum_{i=1}^{M_{max}} \mathds{1}_{\{s_{i} > \alpha \land  \bar{s}_{i} < \alpha\}},
\end{equation}

\begin{equation}
\ell_{LEP} = \sum_{i=1}^{M_{max}} \mathds{1}_{\{ s_{i} < \alpha \land  \bar{s}_{i} > \alpha\}},
\end{equation}
where $\alpha$ is the token threshold, and $\mathds{1}_{\{\cdot\}}$ is the indicator function which equals 1 when the condition inside the braces is true and 0 otherwise. For each stop token, 0 and 1 indicate the mid and the end of a sequence, respectively. 

Combining $\ell_F$ and $\ell_T$, \textit{AllocNet} is therefore trained with the following cumulative loss function
\begin{equation}
    \ell = \ell_F(\bm{t}^*) + w_S \cdot \ell_{T}(\bm{s}),
\end{equation}
where $w_S$ is the weight for balancing $\ell_F$ and $\ell_T$.


\section{Results}
\label{sec:result}

\subsection{Numerical Evaluation}
\subsubsection{Dataset generation} 
We use the real-world flying point cloud dataset M3ED \cite{Chaney_2023_CVPR} and randomly generated maps with Gaussian noise toward obstacles in the simulation for training. 
To model local map regions for online planning, the point cloud data is sampled and cropped into the same size as $12.5 m \times 12.5 m \times 5 m$, re-centered into the same origin. 
\begin{figure}[!ht]
      \centering
      \includegraphics[width=1\columnwidth]{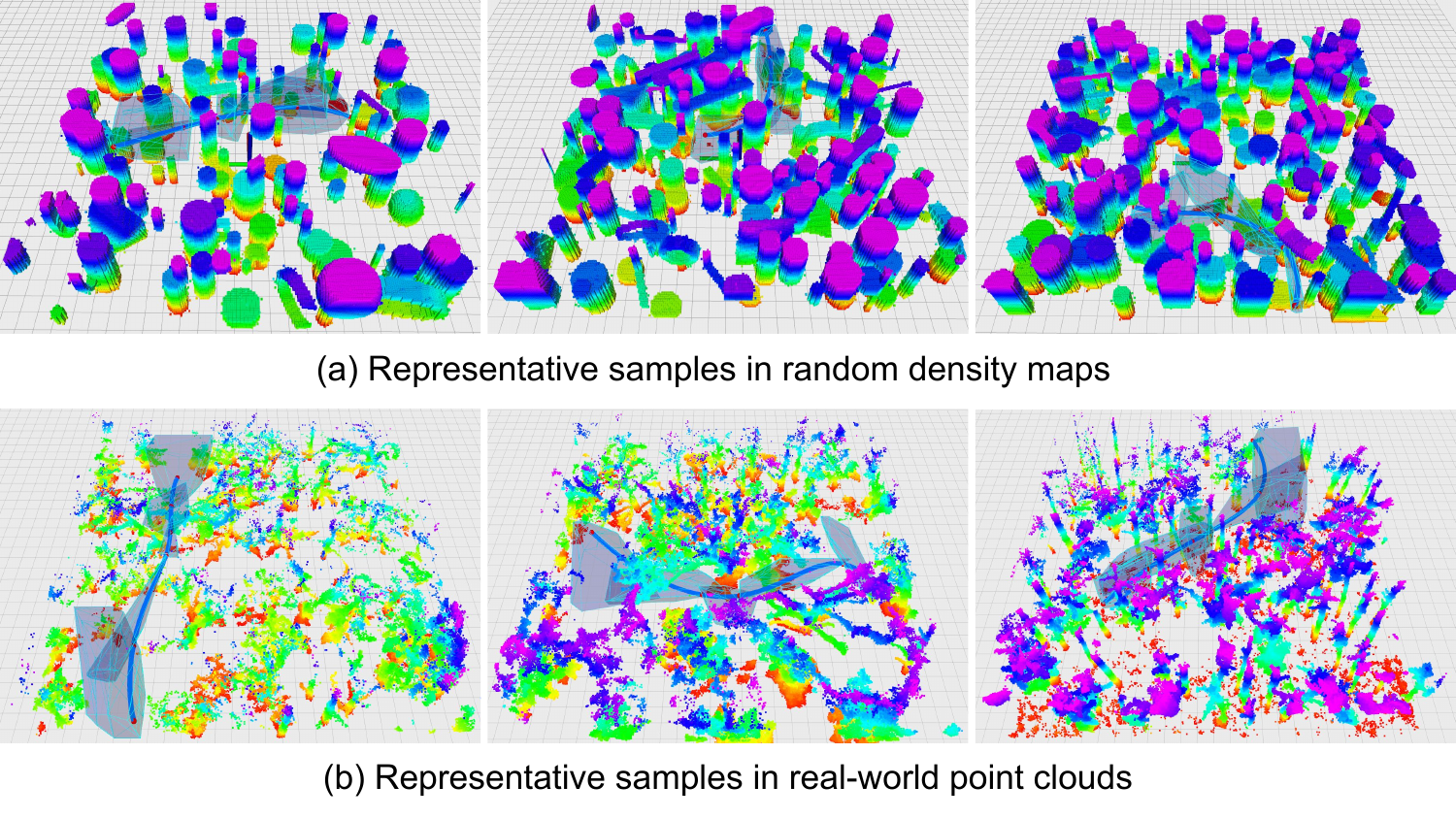}
      \caption{Representative samples in simulation and real-world environments. The simulated map is randomly generated with different types of obstacles and different densities. The real-world flight data are cropped and re-centered. Some sampled paths and corridors are shown in the figure.}
      \label{fig: data_generation}
\end{figure}
We sample 100 valid start and goal states in each map to get 260000 data groups. 
We apply Informed RRT* \cite{6942976} to generate an initial geometric path and use this path to get convex covers as continuous overlapped corridors \cite{7839930}.
\add{Because the runtime increases with the number of hyperplanes, we limit the maximum number of hyperplanes within each corridor to 50. Furthermore, we enforce a maximum trajectory segment of 5 or 10, or this data will be dropped. The training variables and computation 
time will increase with the number of segments based on the problem formulation.}
We use the well-known trajectory optimization methods \cite{9765821} to get the reference feasible time allocation, which enables minimizing the non-fix total time. 
Some representative samples in both types of maps are shown in Fig. \ref{fig: data_generation}.

\subsubsection{Implementation Details}
We use 3-D 5 (or 7)-degree polynomials and solve the quadratic programming via OSQP \cite{osqp}. 
The corridors and flat-based dynamic constraints are approximated discrete hard constraints for designed feasible trajectories. 
We implemented the \textit{AllocNet} in PyTorch and trained it on our generated dataset with a carefully selected set of parameters, shown in Tab. \ref{tab: param}. 
The LSTM component within the network has a hidden size of 256 and comprises a single layer. 
The network is trained on NVIDIA RTX A5000 GPU and later integrated with other ROS modules by using LibTorch.
The model can converge easily in 100 to 500 epochs during training with the reference time loss term. 

\begin{table}[!ht]
    \renewcommand\arraystretch{1.2}
    \centering
    \caption{Parameters\label{tab: param}}
    \begin{tabular}{cccccccc}
      \hline\hline
       $v_{max}$ & $a_{max}$ & $j_{max}$ & $N_{res}$ &$w_F$ & $w_t$ & $w_S$ & $\lambda_{p}$ \\
      \hline
       4 $m/s$ &  6 $m/s^2$  &  8$m/s^3$    & 20 & 1200  &  17.5  &  20  & 5 \\
    \hline\hline
    \end{tabular}
\end{table}

\subsubsection{Ablation Experiment}
We conduct a set of ablation experiments to comprehensively evaluate the performance of  \textit{AllocNet} and gain deeper insights into its mechanisms. 
These experiments involve comparing the learned  \textit{AllocNet} model with versions that include modifications to the loss function and part of the architecture.
To evaluate the role and impact of the individual components and parameters of the proposed model, we prepare several alternative versions of the model:
\begin{itemize}
    \item \textbf{Reference Time} \bm{$\ell_F$} \textbf{Only}: By training a model using only the $\ell_F$ loss function, we aim to investigate the role and contribution of $\ell_T$ to the performance of  \textit{AllocNet}.
    \item \textbf{Reference Time} \bm{$\ell_F$} \textbf{w/ Token}: In this variant, we set the token threshold $\alpha$ as 0.5, which is the most frequently used token threshold for LSTM models in general. This experiment is designed to demonstrate the influence of tuned token thresholds on predicting the correct sequence length.
    \item \textbf{Ours w/ MLP Output Module}: We replace the LSTM output module with a 3-layer MLP with a hidden size of 256. This variant helps us evaluate the advantages of LSTM to predict segment numbers accurately.
    \item \textbf{Ours} \bm{$ (\alpha = 0.35 / 0.75 )$}: We evaluate the model with different stopping token threshold $\alpha$ to investigate the tradeoff of biased threshold to control the actual output length and the final performance.
\end{itemize}
To ensure attributable performance changes to specific modifications, all comparative models maintain the original parameter values for elements that have not been specifically altered for the experiment.
\begin{figure}[!ht]
      \centering
      \includegraphics[width=1\columnwidth]{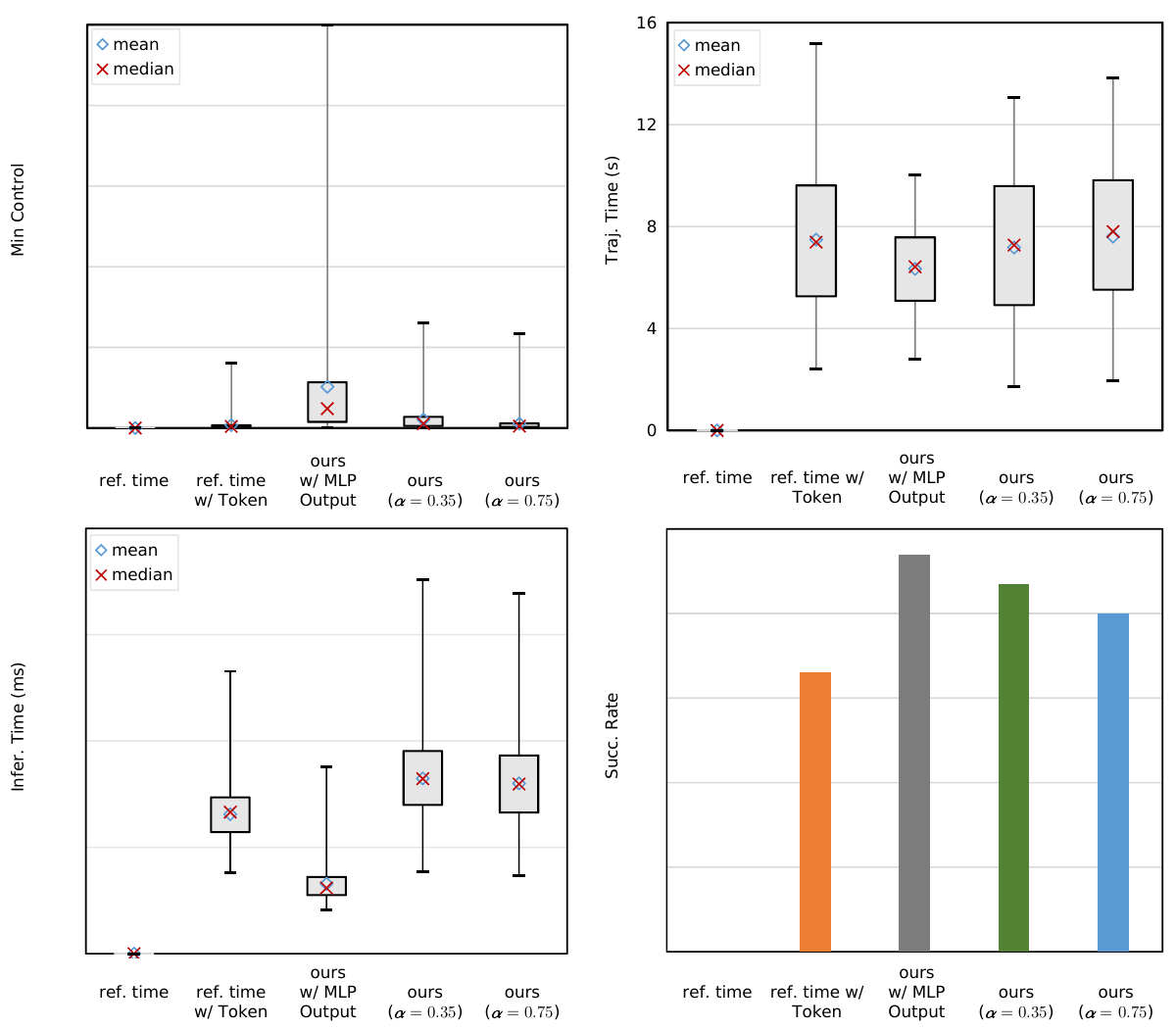}
      \caption{Ablation study on minimum control efforts, total flight time (s), inference time (ms), and success rates. }
      \label{fig: abl}
\end{figure}

The test result is shown in Fig. \ref{fig: abl}. With the use of only objective loss, the value of the output time vector tends to accumulate in the first index and barely works as expected. Adding the token loss significantly alleviates this issue but still stays in highly sub-optimal areas with a larger total time range and lower success rate. 
\add{Although using an MLP layer instead of LSTM shows a higher success rate by consistently outputting the maximum length of values, it lacks precision in predicting the actual length of time sequences due to its non-autoregressive nature. In contrast, the autoregressive nature of LSTM, when trained with token loss, allows for more accurate termination of sequences, leading to lower control efforts compared to MLP, which tends to overfit the most frequent sequence length.}
For our proposed method, the failure scenarios are mainly because the length of valid time allocation doesn't fit the actually input corridor length, which is nontrivial to train together with multiple loss functions. 
The stopping token threshold is usually set as 0.5, but with multiple loss functions, relaxing the threshold ($\alpha$ = 0.35) will accordingly increase the overall success rate. 
Otherwise, using larger threshold results in more conservative total trajectory times and lower control efforts.

\subsection{Benchmark Comparisons in Simulation}

We compare our proposed methods with some existing solving schemes. All methods are provided with dynamic and corridor constraints, with the trajectory represented by piece-wise polynomials. 
Theoretically, using linear inverse with higher derivatives waypoints equality constraints can achieve a more numerically stable solution \cite{7963135}, and our sub-problem formulation can also be replaced by this method. 
To make a fair comparison, we only choose benchmark methods that can be combined with general nonlinear or quadratic programming formulation, and all are formulated with the same general inequality constraints. 
We also observe some learning-based methods with hard constraints \cite{donti_dc3_2021, 9150971}, but the performances are not usually better than the state-of-the-art solver and are even worse for nonlinear problems. 
We evaluate all these methods in four different types of environments: simulated random map, forest, urban, and indoor. 
We utilize only the simulation and forest data during training, whereas the urban and indoor maps come from unseen data distributions. The details of benchmark methods are:

\begin{itemize}
    \item \textbf{QP$^*$} \cite{5980409}: The quadratic programming is solved by OSQP, and the optimal times' gradient descent is solved by backtracking line search using L-BFGS \cite{liu1989limited}. The original problem doesn't enforce dynamic inequality constraints; therefore, it can reach a 100\% success rate. However, we include all these constraints for comparison and will directly exit the line search if it reaches an invalid time allocation. 
    \item \textbf{NLP}: we use NLopt ~\cite{NLopt} to solve the nonlinear programming in Prob. \ref{prob: nonlinear} and use the Improved Stochastic Ranking Evolution Strategy (ISRES) algorithm to get the global optimal solution. Because it's hard to generate the optimal solution fast, we use the QP solution as an initial guess and limit 5 seconds for maximum solving time.
    \item \textbf{Bilevel}~\cite{9147300}: The time allocation and trajectory optimization are combined in a BiLevel Optimization problem. We replace the original trajectory representation as Bezier Curves to polynomials and also change the constraints formulation accordingly. The inner loop quadratic programming is solved by OSQP.
\end{itemize}

\begin{table}[htbp]
    \renewcommand\arraystretch{1.2}
    \caption{Benchmark on Trajectory Optimization\label{tab: benchmark}}
    \begin{tabular}{cl|rrrr}
    \hline 
    & Method & \tabincell{c}{Min\\  Control}  &\tabincell{c}{Traj. \\  Time (s)}   &  \tabincell{c}{Comp. \\  Time (ms)} & \tabincell{c}{Succ. \\  Rate} \\
    \hline
    \multirow{4}*{Sim} &  QP$^*$ & 129.08 & 7.66  & 1820.15 & 0.51 \\
  &NLP &   154.61 &  \textbf{7.00} &  5095.19  &0.79 \\
 & Bilevel OPT & 151.18  & 7.10  &  2030.81&  0.76 \\
    &  Ours ($\alpha $ = 0.35)  &  108.32 & \textbf{7.00} & 55.36 & \textbf{0.84}\\ 
     & Ours ($\alpha $ = 0.75)  &  \textbf{55.14}   &7.54   & \textbf{44.91} & 0.80 \\
    \hline
       \multirow{4}*{Forest} & QP$^*$  &  108.05 & 7.89 & 1619.20 & 0.52 \\
  &NLP  &   138.26 &  7.22 &   5093.21   &0.78 \\
 & Bilevel & 136.79 & 7.29  &  1813.08&  0.77 \\
    &  Ours ($\alpha $ = 0.35)  &  105.21 & \textbf{7.17} & 55.48 & \textbf{0.87} \\ 
     & Ours ($\alpha $ = 0.75)  & \textbf{51.75} & 7.66 & \textbf{41.56} & 0.79 \\
      \hline
  \multirow{4}*{Urban} &  QP$^*$  & 104.86 & 8.05 & 1594.43   & 0.51\\
 &NLP  & 129.62  & \textbf{7.22} & 5095.45   & 0.81\\
 & Bilevel   &124.32 & 7.40 & 1835.72   & 0.79 \\
 &  Ours ($\alpha $ = 0.35)  & 101.03 & 7.23 & 54.39   & \textbf{0.86}\\ 
 & Ours ($\alpha $ = 0.75)   & \textbf{54.97} & 7.63  & \textbf{43.29}  &  0.80\\
    \hline
     \multirow{3}*{Indoor}   &  QP$^*$  &  104.63 & 8.02 & 1481.62 & 0.52\\
  &NLP&  130.90 & 7.32 & 5089.54 & 0.80 \\
 & Bilevel  & 127.06 & 7.43 & 1769.22 & 0.78 \\
    &  Ours ($\alpha $ = 0.35)  &  102.64      &  \textbf{7.34} & 52.23 &   \textbf{0.90} \\ 
     & Ours ($\alpha $ = 0.75)  &  \textbf{56.09} & 7.74 & \textbf{42.31}  &0.79 \\
    \hline
    \end{tabular}
\end{table}

As indicated in Tab. \ref{tab: benchmark}, we apply a traditional method to generate the safe flight corridor for all different datasets, ensuring there are no generalization issues in our proposed framework and comparison. 
In addition, we set the same relative tolerance criteria and limit the iterations for all the solvers.
The NLP methods usually cannot converge with the allowed time, and the performance could be worse if no feasible initial guess is provided. 
The QP$^*$ and Bilevel methods need an initial total time, whereas other methods can directly optimize the time during optimization.
The QP$^*$ method uses finite differences as the gradient direction and has a relatively lower success rate among all other methods. This is due to the fact that the quadratic programming will easily become infeasible because of invalid time allocation. 
The benchmark results highlight the advantages in terms of both computation time and also minimum control efforts of our proposed method. 
Even when the total solving time for constrained quadratic programming is factored in, our proposed method can still achieve fast computation time and a reasonable success rate for online planning. 
For online planning, we can combine multiple models to get better overall performance in practice.

\begin{figure}[!ht]
    \centering
    \includegraphics[width=1\columnwidth]{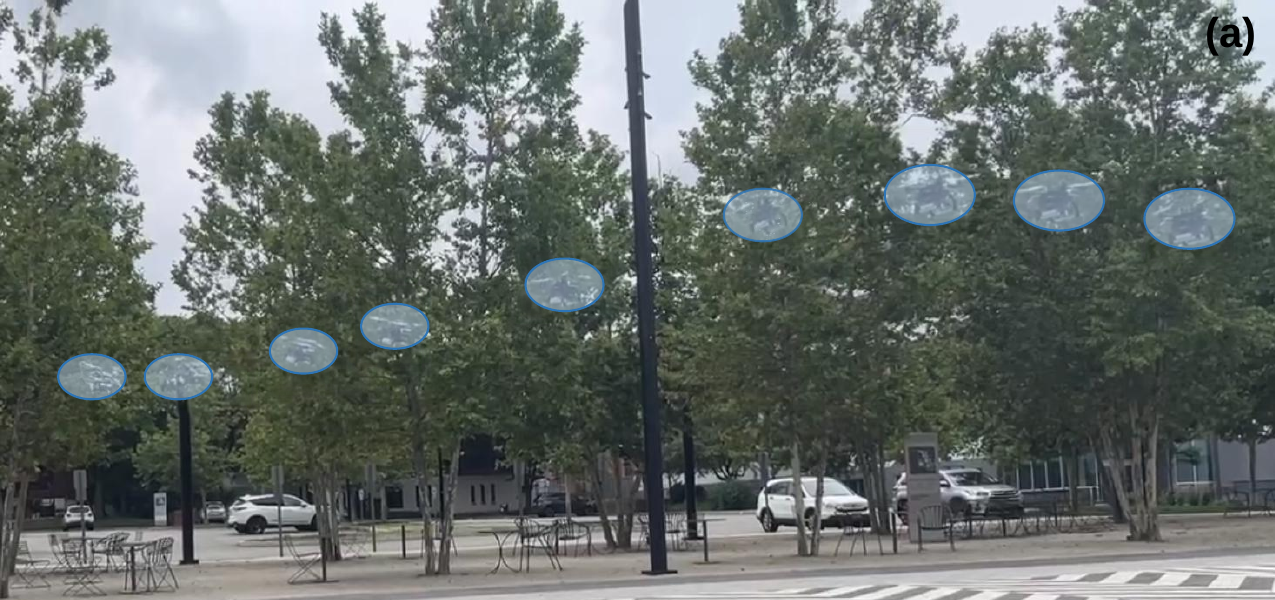}
     \\[\medskipamount]
    \includegraphics[width=1\columnwidth]{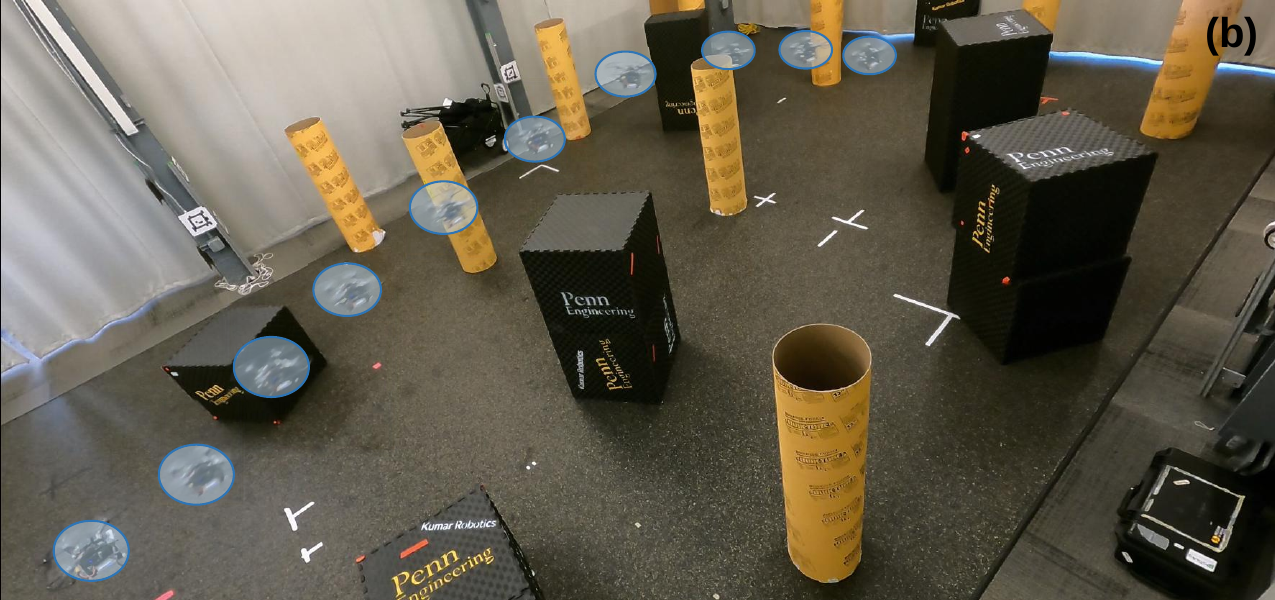}
      \caption{Snapshots of flight experiments in the outdoor (a) and indoor (b) cluttered environments.}
      \label{fig: exp}
\end{figure}

\subsection{Real-world Deployment}

We validate the proposed method on our custom-designed Falcon 250 v2 platform \cite{10160295}, with ModalAI VOXL board\footnote{https://www.modalai.com/pages/voxl} and Intel Realsense camera D435i for real-time state estimation and volumetric mapping. 
The inference runs around 1.0 ms on an i7-10710U CPU, which is similar to the computation time achieved when running on both CPU and GPU versions.
Given the limited sensor range, the box-bounded maximum velocity and acceleration are set to 2.0 $m/s$  to 3.0 $m/s^2$ respectively. 
We utilize a finite horizon local planning style with a 100 HZ finite state machine to call the trajectory optimization module.
When calling the replanning with nonhomogeneous boundary conditions, we also employ models trained from non-rest to rest states models to handle these scenarios. \add{Since} we use simple geometrical pathfinding to generate corridors, the performance would be better if using some dynamic planning methods to get dynamically feasible corridors. 
We extensively demonstrate our framework in both indoor and outdoor cluttered environments, and the snapshots are shown in Fig. \ref{fig: exp}.

\section{Conclusion}

In this paper, we introduce an interpretable and lightweight learning-based planning framework that utilizes neural networks to learn the optimal time allocation for constrained trajectory optimization problems. 
We evaluate our proposed method through extensive simulations and indoor and outdoor experiments demonstrating its efficacy for real-time safe operation. 
Our proposed method relies on the efficiency of the quadratic programming solver and the quality of the initial path. The \add{latter} forces the solution to lie in the same homotopy class as the initial path \add{for corridor generation}. Our future work will focus on integrating efficient and optimal search-based path initialization and corridor generation on GPUs.

\section{Acknowledgments}

We would like to thank Fernando Cladera, Yuezhan Tao, Yifei (Simon) Shao, and Alex Zhou for helping with the results in the paper. The data sets used here are described in \cite{Chaney_2023_CVPR}. The experimental platform was developed for the work presented in \cite{10160295}. 

\bibliography{ref}

\end{document}